\documentclass[10pt,twocolumn,letterpaper]{article}

\usepackage[pagenumbers]{style} % To force page numbers, e.g. for an arXiv version

\usepackage{graphicx}
\usepackage{amsmath}
\usepackage{amssymb}
\usepackage{booktabs}
\usepackage{makecell}
\usepackage{enumitem}
\usepackage{wrapfig}
\usepackage{caption}
\usepackage{marvosym}
\usepackage{tablefootnote}
\usepackage{multirow}
\usepackage{caption}
\usepackage{subcaption}
\usepackage[table]{xcolor}
\usepackage{adjustbox}
\usepackage{tabularx}

\usepackage[pagebackref,breaklinks,colorlinks]{hyperref}

\usepackage[capitalize]{cleveref}
\crefname{section}{Sec.}{Secs.}
\Crefname{section}{Section}{Sections}
\Crefname{table}{Table}{Tables}
\crefname{table}{Tab.}{Tabs.}

\def\name{SiameseIM}
\def\eg{\textit{e.g.,~}}
\def\ie{\textit{i.e.,~}}

\definecolor{mygray}{gray}{0.7}
\newcommand{\demph}[1]{\textcolor{mygray}{#1}}
\definecolor{defaultcolor}{gray}{.93}
\newcommand{\default}[1]{\cellcolor{defaultcolor}{#1}}

\definecolor{gain}{HTML}{34a853}  %
\newcommand{\gain}[1]{\textcolor{gain}{#1}}
\definecolor{lost}{HTML}{ea4335}  %

\newcommand{\res}[2]{{#1} {({\gain{#2}})}}

\newcommand\blfootnote[1]{%
\begingroup
\renewcommand\thefootnote{}\footnote{#1}%
\addtocounter{footnote}{-1}%
\endgroup
}

\begin{document}

\title{Siamese Image Modeling for Self-Supervised \\ Vision Representation Learning}

\author{%
  Chenxin Tao$^{1*\dag}$,
  Xizhou Zhu$^{2*}$,
  Weijie Su$^{3*\dag}$,
  Gao Huang$^{1}$,
  Bin Li$^{3}$,\\
  Jie Zhou$^{1}$,
  Yu Qiao$^{4}$,
  Xiaogang Wang$^{5}$,
  Jifeng Dai$^{1,4}$\textsuperscript{\Letter} \\
$^1$Tsinghua University, $^2$SenseTime Research, $^3$University of Science and Technology of China,\\
$^4$Shanghai Artificial Intelligence Laboratory, $^{5}$The Chinese University of Hong Kong, \\
{\tt\small tcx20@mails.tsinghua.edu.cn, zhuwalter@sensetime.com}, \\
{\tt\small jackroos@mail.ustc.edu.cn, \{gaohuang,jzhou,daijifeng\}@tsinghua.edu.cn,}\\
{\tt\small binli@ustc.edu.cn, qiaoyu@pjlab.org.cn, xgwang@ee.cuhk.edu.hk}}
\maketitle

\begin{abstract}
\vspace{-0.4cm}
Self-supervised learning (SSL) has delivered superior performance on a variety of downstream vision tasks. Two main-stream SSL frameworks have been proposed, i.e., Instance Discrimination (ID) and Masked Image Modeling (MIM). ID pulls together representations from different views of the same image, while avoiding feature collapse. It lacks spatial sensitivity, which requires modeling the local structure within each image. On the other hand, MIM reconstructs the original content given a masked image. It instead does not have good semantic alignment, which requires projecting semantically similar views into nearby representations. To address this dilemma, we observe that (1) semantic alignment can be achieved by matching different image views with strong augmentations; (2) spatial sensitivity can benefit from predicting dense representations with masked images. Driven by these analysis, we propose Siamese Image Modeling (SiameseIM), which predicts the dense representations of an augmented view, based on another masked view from the same image but with different augmentations. SiameseIM uses a Siamese network with two branches. The online branch encodes the first view, and predicts the second view's representation according to the relative positions between these two views. The target branch produces the target by encoding the second view. SiameseIM can surpass both ID and MIM on a wide range of downstream tasks, including ImageNet finetuning and linear probing, COCO and LVIS detection, and ADE20k semantic segmentation. The improvement is more significant in few-shot, long-tail and robustness-concerned scenarios. Code shall be released at {\small\url{https://github.com/fundamentalvision/Siamese-Image-Modeling}}.

\end{abstract}
\vspace{-1.0em}
\blfootnote{\noindent $^{*}$Equal contribution. $^{\dag}$This work is done when Chenxin Tao and Weijie Su are interns at Shanghai Artificial Intelligence Laboratory. \textsuperscript{\Letter}Corresponding author.}
\section{Introduction}
\label{sec:intro}

Self-supervised learning (SSL) has been pursued in the vision domain for a long time~\cite{jing2020self}. It enables us to pre-train models without human-annotated labels, which makes it possible to exploit huge amounts of unlabeled data.
SSL has provided competitive results against supervised pre-training baselines in various downstream tasks, including image classification~\cite{bao2021beit, he2021masked}, object detection~\cite{li2021benchmarking} and semantic segmentation~\cite{he2020momentum}.

To effectively train models in the SSL manner, researchers design the so-called ``pretext tasks'' to generate supervision signals. One of the most typical frameworks is \textit{Instance Discrimination (ID)}, whose core idea is to pull together representations of different augmented views from the same image, and avoid representational collapse. Different variants of ID have been proposed, including contrastive learning~\cite{he2020momentum,chen2020simple}, asymmetric networks~\cite{grill2020bootstrap,chen2021exploring}, and feature decorrelation~\cite{zbontar2021barlow,bardes2021vicreg}. A recent work~\cite{tao2021exploring} has shown the intrinsic consistency among these methods via their similar gradient structures. For ID methods, the representations of each image are well separated, thus inducing good linear separability. However, as shown in \cite{li2021benchmarking}, for transfer learning on detection tasks with Vision Transformers~\cite{dosovitskiy2020image}, ID is not superior to supervised pre-training, and even lags behind random initialization given enough training time.

Recently, another SSL framework has gradually attracted more attention, namely \textit{Masked Image Modeling (MIM)}~\cite{bao2021beit,he2021masked}. MIM methods train the model to reconstruct the original content from a masked image. Such practice can help to learn the rich local structures within an image, leading to excellent performance in dense prediction tasks such as object detection~\cite{li2021benchmarking}. Nevertheless, MIM does not have good linear separability as ID, and usually performs poorly under the few-shot classification settings~\cite{assran2022masked}.

\setlength{\tabcolsep}{4pt}
\renewcommand{\arraystretch}{1.1}
\begin{table*}[t]
    \small
    \centering
    \resizebox{0.75\linewidth}{!}{
    \begin{tabular}{l|ccc|cc|c|cc|c}
    \toprule
         & \multicolumn{3}{c|}{ImageNet} & \multicolumn{2}{c|}{COCO} & ADE20k & \multicolumn{2}{c|}{LVIS} & Robustness \\
         & FT & LIN & FT$_{1\%}$ & AP$^\mathrm{b}$ & AP$^\mathrm{m}$ & mIoU & AP$^\mathrm{b}_\mathrm{rare}$ & AP$^\mathrm{m}_\mathrm{rare}$ & avg score$^*$ \\
    \midrule
        MoCo-v3 (ID method) & 83.0 & 76.7 & 63.4 & 47.9 & 42.7 & 47.3 & 25.5 & 25.8 & 43.4 \\
        MAE (MIM method) & 83.6 & 68.0 & 51.1 & 51.6 & 45.9 & 48.1 & 29.3 & 29.1 & 41.8 \\
    \midrule
        SiameseIM (ours) & 84.1 & 78.0 & 65.1 & 52.1 & 46.2 & 51.1 & 30.9 & 30.1 & 47.9\\
        Improve w.r.t. MoCo-v3 & \gain{+1.1} & \gain{+1.3} & \gain{+1.7} & \gain{+4.2} & \gain{+3.5} & \gain{+3.8} & \gain{+5.4} & \gain{+4.3} & \gain{+4.5} \\
        Improve w.r.t. MAE & \gain{+0.5} & \gain{+10.0} & \gain{+14.0} & \gain{+0.5} & \gain{+0.3} & \gain{+3.0} & \gain{+1.6} & \gain{+1.0} & \gain{+6.1}\\
    \bottomrule
    \end{tabular}}
    \vspace{-0.5em}
    \caption{\name\ surpasses MoCo-v3 (ID method) and MAE (MIM method) on a wide range of downstream tasks. $^*$The robustness average score is calculated by averaging top-1 acc of IN-A, IN-R, IN-S, and 1-mCE of IN-C. For detailed results, please refer to Section~\ref{subsec:main results}.}
    \label{tab:intro-results}
\end{table*}

\begin{figure*}[t]
    \centering
    \includegraphics[width=0.95\textwidth]{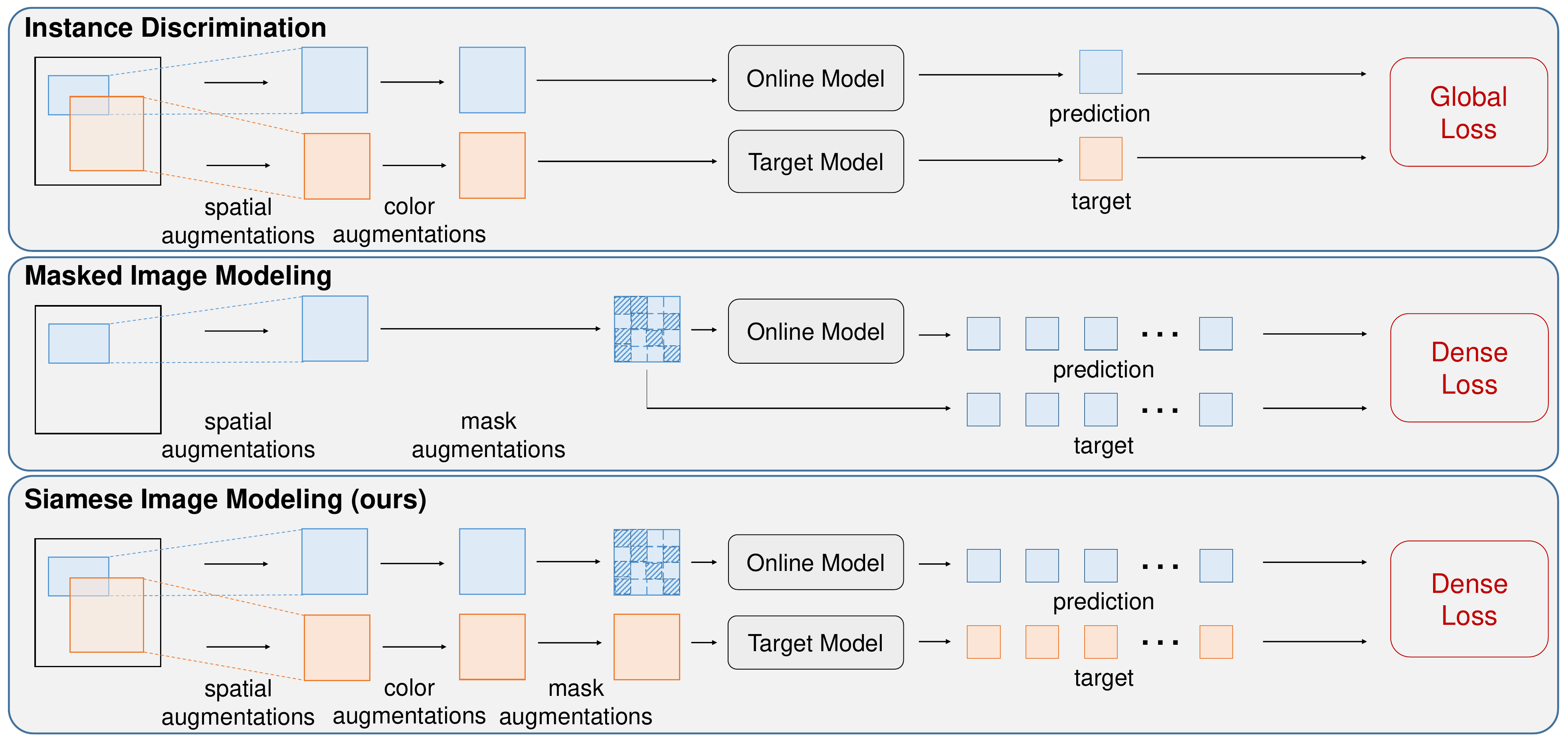}
    \vspace{-0.5em}
    \caption{Comparisons among ID, MIM and \name{}. Matching different augmented views can help to learn semantic alignment, which is adopted by ID and \name{}. Predicting dense representations from masked images is beneficial to obtain spatial sensitivity, which is adopted by MIM and \name{}.}
    \label{fig:compare}
    \vspace{-1.5em}
\end{figure*}

Both ID and MIM methods have their own strengths and weaknesses. We argue that this dilemma is caused by neglecting the representation requirements of either semantic alignment or spatial sensitivity. Specifically, MIM operates within each image independently, regardless of the inter-image relationship. The representations of semantically similar images are not well aligned, which further results in poor linear probing and few-shot learning performances of MIM. On the other hand, ID only uses a global representation for the whole image, and thus fails to model the intra-image structure. The spatial sensitivity of features is therefore missing, and ID methods usually produce inferior results on dense prediction.

To overcome this dilemma, we observe the key factors for semantic alignment and spatial sensitivity:
(1) semantic alignment requires that images with similar semantics are projected into nearby representations. This can be achieved by matching different augmented views from the same image. Strong augmentations are also beneficial because they provide more invariance to the model; (2) spatial sensitivity needs modeling the local structures within an image. Predicting dense representations from masked images thus helps, because it models the conditional distribution of image content within each image. These observations motivate us to predict the dense representations of an image from a masked view with different augmentations.

To this end, we propose Siamese Image Modeling (\name{}), which reconstructs the dense representations of an augmented view, based on another masked view from the same image but with different augmentations (see Fig.~\ref{fig:compare}). It adopts a Siamese network with an online and a target branch. The online branch consists of an encoder that maps the first masked view into latent representations, and a decoder that reconstructs the representations of the second view according to the relative positions between these two views. The target branch only contains a momentum encoder that encodes the second view into the prediction target. The encoder is made up of a backbone and a projector. After the pre-training, we only use the online backbone for downstream tasks.

As shown in Tab.~\ref{tab:intro-results}, \name\ is able to surpass both MIM and ID methods over a wide range of evaluation tasks, including full-data fine-tuning, few-shot learning and linear probing on ImageNet~\cite{deng2009imagenet}, object detection on COCO~\cite{lin2014microsoft} and LVIS~\cite{gupta2019lvis}, semantic segmentation on ADE20k~\cite{zhou2017scene}, as well as several robustness benchmarks~\cite{hendrycks2021natural,hendrycks2021many,hendrycks2019benchmarking,wang2019learning}. By gathering semantic alignment and spatial sensitivity in one model, \name\ can deliver superior results for all tasks.
We also note that such improvements are more obvious on ADE20k($\sim$3 points) and LVIS($\sim$1.6 point for rare classes) datasets. These long-tailed datasets demands semantic alignment and spatial sensitivity at the same time, and \name\ thus can deliver superior performance on them.

Our contributions can be summarized as follows:

\begin{itemize}[leftmargin=1em]
    \vspace{-0.7em}
    \item As a new form of SSL, \name\ is proposed to explore the possibilities of self-supervised pre-training. It displays for the first time that that using only a single dense loss is enough to learn semantic alignment and spatial sensitivity well at the same time;
    \vspace{-0.7em}
    \item Compared with MIM methods, \name\ shows that reconstructing another view helps to obtain good semantic alignment. This also suggests that MIM framework can be used to reconstruct other targets with proper guidance, which opens a possible direction for MIM pretraining;
    \vspace{-0.7em}
    \item Compared with ID methods, \name\ shows that dense supervision can be applied by matching the dense correspondence between two views strictly through their relative positions. We demonstrate dense supervision can bring a considerable improvement of spatial sensitivity;
    \vspace{-0.7em}
    \item \name\ is able to surpass both MIM and ID methods over a wide range of tasks. \name\ obtains more improvements in few-shot, long-tail and robustness-concerned scenarios.
\end{itemize}

\section{Related work}
\label{sec:related}

\noindent\textbf{Instance Discrimination (ID).~} The core idea of instance discrimination is to pull together different augmented views of the same image and avoid representational collapse~\cite{wu2018unsupervised,van2018representation}. In this way, models can learn to separate the representation of each image, leading to decent linear separability.
There are three typical types of instance discrimination methods, while Siamese networks are always employed. 
\textit{Contrastive Learning} methods~\cite{he2020momentum, chen2020improved, chen2021empirical, chen2020simple} push apart views from different images (negative samples) to avoid representational collapse.
\textit{Asymmetric Network} methods~\cite{grill2020bootstrap, chen2021exploring} 
explore to get rid of negative samples with the help of an asymmetric network design. In these methods, a predictor network is only appended after one branch of the siamese network, and the other branch is detached from the gradient back-propagation. 
\textit{Feature Decorrelation} methods~\cite{zbontar2021barlow, bardes2021vicreg, hua2021feature} try to accomplish instance discrimination by reducing the redundancy among different feature dimensions.
These different methods are then unified in UniGrad~\cite{tao2021exploring} by revealing that they share similar gradient structures.
A recent work~\cite{tomasev2022pushing} has even successfully surpassed the performance of supervised learning with ResNets~\cite{he2016deep}. There have also been works on applying instance discrimination to Vision Transformers~\cite{chen2021empirical,caron2021emerging}, which demonstrate impressive performances.

The most common evaluation metric used for ID is linear probing, which trains a linear classifier on top of frozen representations. This metric concentrates on the linear separability of learned features. However, as \cite{he2021masked} has pointed out, the dense prediction performance of ID on Vision Transformers is not superior to supervised pre-training, especially on object detection tasks.

Some previous ID works~\cite{liu2020self,wang2021dense, xie2021propagate, li2021efficient,wei2021aligning,xiong2021self} have tried to introduce dense supervision to enhance local features. They employ different techniques to build dense correspondence between two views, including using Earth Mover's Distance~\cite{liu2020self}, matching feature similarity~\cite{wang2021dense,li2021efficient}, finding nearest neighbors~\cite{xie2021propagate}, using extra region proposal or flow modules~\cite{xiong2021self}.
However, these works either only focus on detection result, or rely on an extra global loss to improve linear probing result, which still requires to find a trade-off between semantic alignment and spatial sensitivity.

Our work display for the first time that only using a dense loss is enough to learn these two properties well at the same time.
Unlike previous methods, we utilizes the relative positions between two views to strictly align the spatial correspondence. In doing so, \name\ outperforms the original ID methods by a large margin on a strong detection baseline.

\vspace{0.5em}
\noindent\textbf{Masked Image Modeling (MIM).~} Masked image modeling intends to reconstruct image content from a masked image, which is motivated by the masked language modeling in NLP~\cite{devlin2018bert,radford2018improving,radford2019language,brown2020language}.
iGPT~\cite{chen2020generative} first tries to reconstruct image pixels. ViT~\cite{dosovitskiy2020image} has also tried to predict the mean color of the masked patch. However, these preliminary attempts are not competitive with their supervised counterparts. BEiT~\cite{bao2021beit} reveals the power of MIM by predicting visual tokens from a pre-trained discrete VAE~\cite{ramesh2021zero}. MAE~\cite{he2021masked} successfully performs pre-training via predicting raw pixels. It shows that the key point is to use a high masked ratio due to the high spatial redundancy. After that, different works~\cite{dong2021peco,wei2021masked,el2021large,baevski2022data2vec,chen2022context} continue to push the limit by improving the quality of prediction targets. Some works~\cite{wei2021masked,baevski2022data2vec} have shown that it is more effective to predict features rather than raw pixels for learning representations.

Unlike ID methods, MIM methods excel in transfer learning with full model fine-tuning with Vision Transformers, but lack good linearly-separated representations~\cite{he2021masked}. For example, given enough training epochs, BEiT~\cite{bao2021beit} and MAE~\cite{he2021masked} can surpass other pre-training paradigms on detection tasks. However, under few-shot scenes, MIM methods are not as data-efficient as ID methods because of their poor linear separability~\cite{assran2022masked}.

\begin{figure*}[t]
    \centering
    \includegraphics[width=0.9\textwidth]{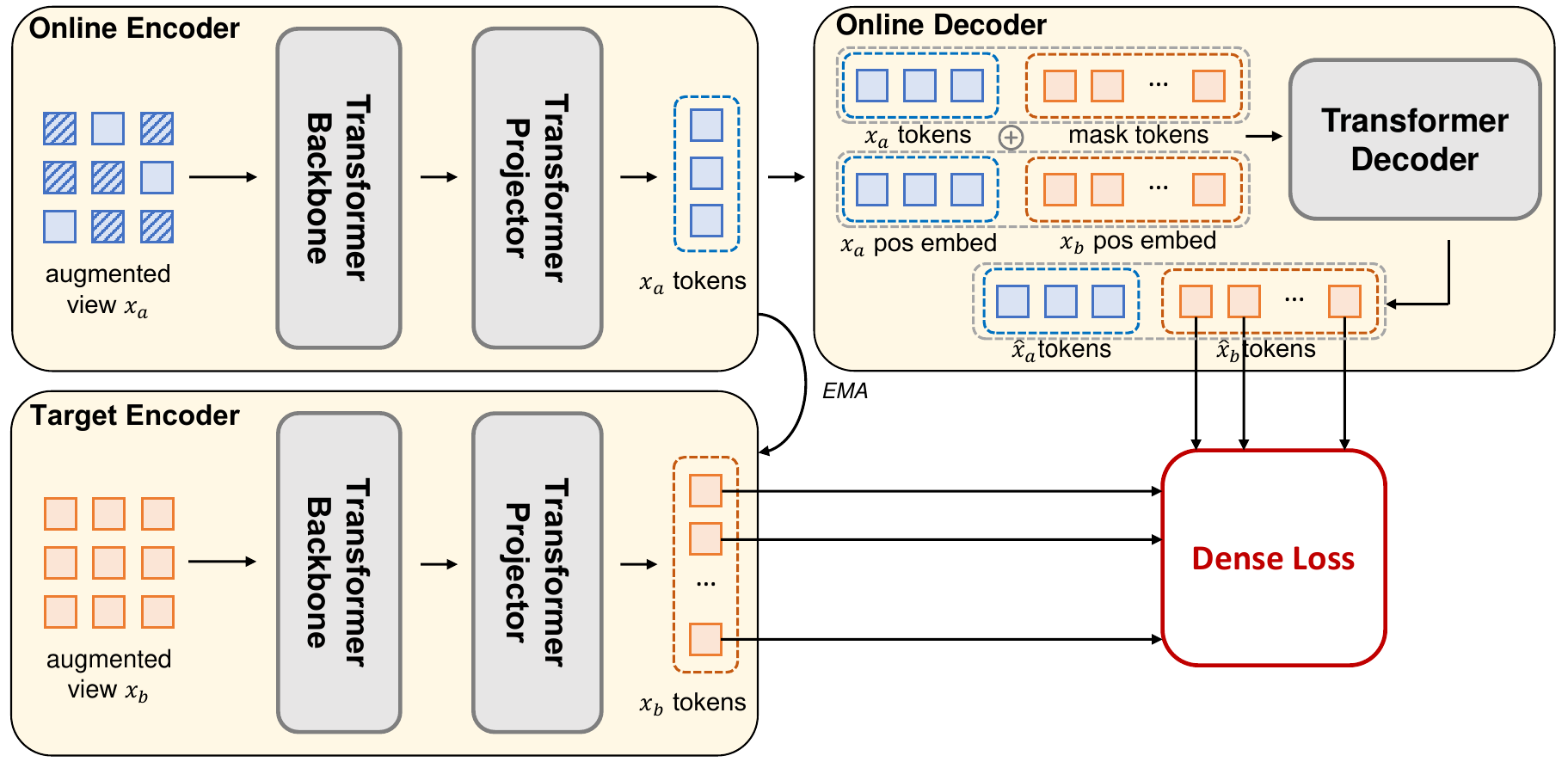}
    \vspace{-0.1cm}
    \caption{The overview of our Siamese Image Modeling (\name{}). Different augmented views are fed into the online and target branches. The online encoder operates on the visible patches of $x_a$. The online decoder accepts the $x_a$ tokens as well as mask tokens that correspond to predicted $x_b$ tokens. We use relative positions to inform the locations between $x_a$ and $x_b$. The target encoder maps $x_b$ to the target representations. We finally apply the dense loss on the dense representations.}
    \label{fig:overview}
    \vspace{-0.7em}
\end{figure*}

Our work demonstrates that MIM can also produce the same adequate linear separable representations as ID. This is achieved by predicting the representation of another augmented view from the same image, rather than reconstructing the original view. Through reconstructing another view, \name\ can even surpass the linear probing performance of ID methods.
\section{Method}
\label{sec:method}
We depict our model of \name\ in Fig.~\ref{fig:overview}. It takes two augmented views $x_a$ and $x_b$ of the same image as inputs. \name\ aims to predict the dense representations of $x_b$ based on that of $x_a$. A Siamese network with an online and a target branch is used. The online branch is made up of an encoder that encodes the visible patches of $x_a$ into a latent representation, and a decoder that predicts the representation of $x_b$ according to the relative positions between $x_a$ and $x_b$. The target branch only has a momentum encoder which takes $x_b$ as input. The encoder consists of a backbone and a projector. After the pre-training, only the online backbone is used for downstream evaluation.

%%%%%%%%%%%%%%%%%%%%%%%%%%%%%%%%%%%%%%%%%%%%%%%%%%%%%%%%%%%%%%%%%%%%%%%%%%%%%%%%%%%%%%%%%%%%%%%%%%%%%%%%%%%%%%%%%%%%%%%%%%%%%%%%%%%%%%%%%%%%%%%%%%

\subsection{Augmented Inputs}
ID methods adopt two different augmented views, while MIM methods utilize a single view. In our method, similar to ID methods, we feed two different views as the inputs to the online and target branches, respectively. As will be shown in Section~\ref{subsec:ablation}, different views can significantly increase the linear probing result without harming the performance of object detection.

Apart from the number of views, there are also differences in the augmentations of previous methods (see Fig.~\ref{fig:compare}). ID methods~\cite{he2020momentum,chen2020simple,chen2021empirical} tends to add stronger augmentations, which typically contain spatial and color augmentations. Whereas recently, MAE~\cite{he2021masked} reports that color augmentations are not beneficial for MIM pre-training. We find that color augmentations have different effects under different training settings. They can provide more invariance when used with different views, but such effect will vanish if paired with the same view (more analysis can be found in Section~\ref{subsec:ablation}). As a result, we reserve both the spatial and color augmentations from ID methods~\cite{chen2021empirical}.

Another difference is that MIM masks out some patches of the input image for reconstruction, which we refer as mask augmentation. With mask augmentation, the task of dense prediction can model the conditional distribution of image content within each image. The representations are trained to capture the local structure, and thus are endowed with spatial sensitivity. Therefore, we also apply mask augmentation to the view of the online branch.

%%%%%%%%%%%%%%%%%%%%%%%%%%%%%%%%%%%%%%%%%%%%%%%%%%%%%%%%%%%%%%%%%%%%%%%%%%%%%%%%%%%%%%%%%%%%%%%%%%%%%%%%%%%%%%%%%%%%%%%%%%%%%%%%%%%%%%%%%%%%%%%%%%

\subsection{Prediction Targets}
There can be multiple choices for the prediction targets. For example, ID methods select to predict the features of different augmented views, while MIM methods are designed to predict pixels or features of the same view. We empirically find that feature prediction is superior for different views (see Section~\ref{subsec:ablation}). \name\ is thus designed to predict the features of another different augmented view from the same image. We shall describe how the prediction and target are calculated.% below, as well as why such design is chosen.

\vspace{0.5em}
\noindent\textbf{Online Branch} will make the prediction. The online encoder first maps the masked view $x_a$ into a latent representation $y_a \in \mathbb{R}^{N_v\times D}$, where $N_v$ denotes the number of visible patches and $D$ is the feature dimension. Following the practice in ID methods~\cite{chen2020simple,wang2021revisiting}, we append a projector after the backbone to form the encoder. The online decoder $g(\cdot)$ then combines $y_a$, mask tokens $m$, and their relative positions to calculate the prediction $y_b \in \mathbb{R}^{N\times D}$ as
\begin{equation}
    y_b = g\bigg(\mathrm{Concat}\left(y_a+p_a,~ \{m+p_b^{(u,v)}\}_{u=1,v=1}^{N_h,N_w}\right)\bigg).
\end{equation}
Here, $m$ indicates the learnable embedding of the mask token, which follows \cite{he2021masked}. $p_a$ is the position embeddings for $x_a$, and $p_b^{(u,v)}$ denotes the positional embedding for the patch of $x_b$ at location $(u, v)$, which will be introduced later. $N_h$ and $N_w$ denote the number of tokens along height and width dimensions in $x_b$ (\eg $N_h = N_w = 14$), respectively. $N=N_h\times N_w$ is therefore the number of all tokens in $x_b$. Note that different from MIM methods, the mask tokens correspond to image patches from the different target view $x_b$ instead of the input view $x_a$.

\vspace{0.5em}
\noindent\textbf{Target Branch} is responsible for producing the target. The target encoder is an exponential moving average of the online encoder. It takes all tokens from $x_b$ as input and outputs their latent representation $z_b \in \mathbb{R}^{N\times D}$. Note that we can also directly predict the raw pixels, where the target encoder is unnecessary. However, we find that feature prediction can give better performance when different views are used (see Section~\ref{subsec:ablation}).

\begin{figure}
    \centering
    \includegraphics[width=0.98\linewidth]{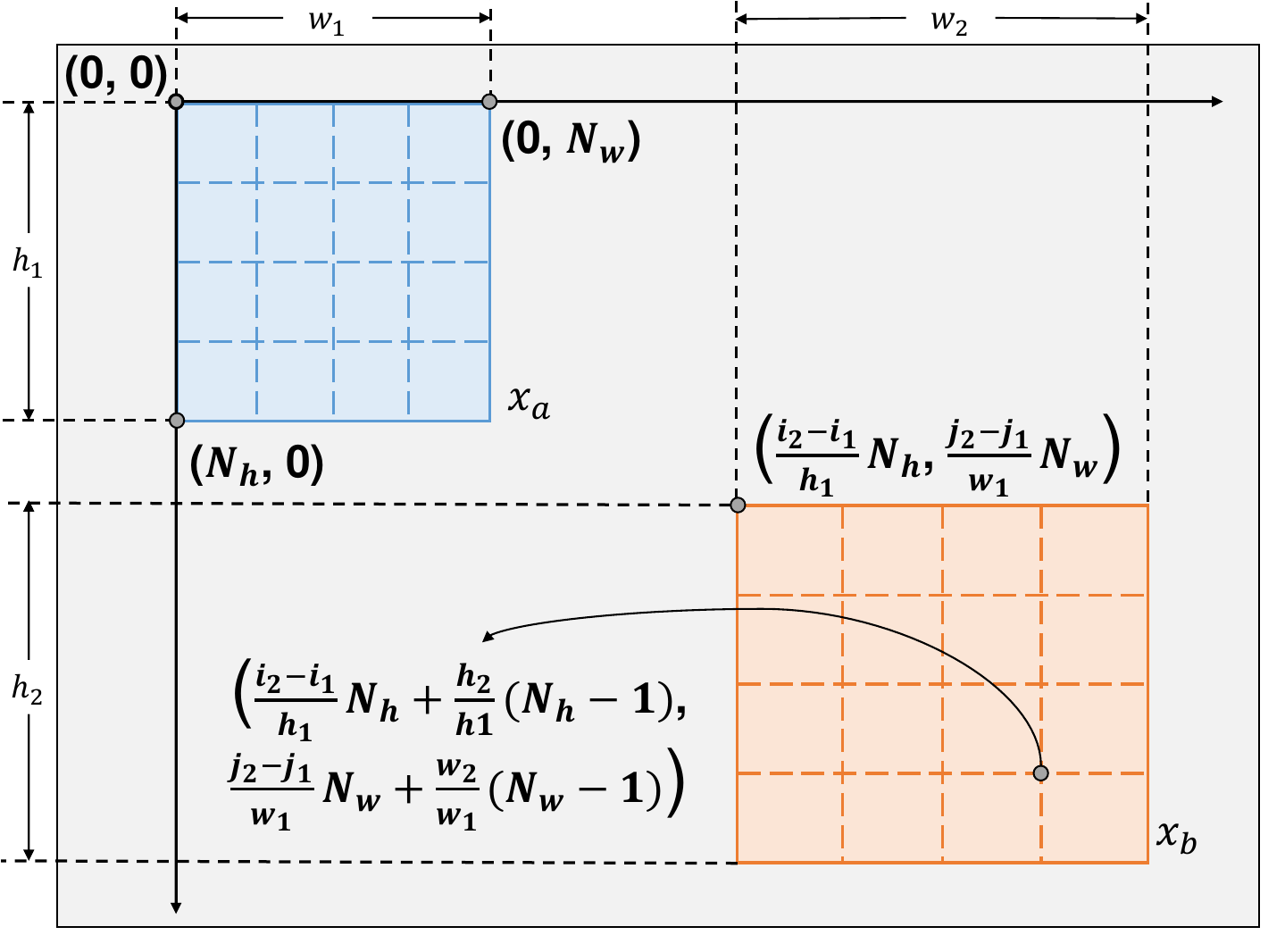}
    \vspace{-0.5em}
    \caption{Positional embedding for online decoder. The positions are calculated with respect to the left-top origin of $x_a$.}
    \vspace{-0.5em}
    \label{fig:pos_embed}
    \vspace{-0.5em}
\end{figure}

\vspace{0.5em}
\noindent\textbf{Positional Embedding for Online Decoder} is necessarily required to inform the decoder of the corresponding locations of each patch in $x_a$ and $x_b$. The decoder predicts the dense representations of $x_b$ based on visible patches from $x_a$ and their corresponding locations. 
For all input patches to the online decoder, including visible patches from $x_a$ and mask tokens indicating $x_b$, their positional embeddings are calculated from the relative position with respect to the left-top origin of $x_a$. Fig.~\ref{fig:pos_embed} shows the detailed process.
Suppose the (left, top, height, width)  positional properties of the two cropped views $x_a$ and $x_b$ in the original image are $(i_1, j_1, h_1, w_1)$ and $(i_2, j_2, h_2, w_2)$, respectively.
The positions for $x_a$ and mask tokens indicating $x_b$ are
\begin{align}
    \tilde{p}_a^{(u,v)} &= (u-1, v-1), \nonumber\\
    \tilde{p}_b^{(u,v)} &= \adjustbox{max width=0.75\linewidth}{$\bigg(\frac{h_2}{h_1}(u-1) + \frac{i_2-i_1}{h_1}N_h, \frac{w_2}{w_1}(v-1) + \frac{j_2-j_1}{w_1}N_w\bigg),$}
\end{align}
where $u$ and $v$ are the location indexes along height and width dimensions.
We further apply $\sin(\cdot)$ and $\cos(\cdot)$ operators to get the 2-D sin-cos positional embeddings, following the practice in MAE~\cite{he2021masked}. If the same view is used for input and target, $\tilde{p}_b^{(u,v)}$ will degenerate to $\tilde{p}_a^{(u,v)}$ as used in MAE. 
Instead, we choose to use different views because they are crucial for improving the linear separability of the representation (see Section~\ref{subsec:ablation}). Moreover, to inform the scale variation between two different views, we add the relative scale changes as $s=\left(10\log\frac{h_2}{h_1},\quad 10\log\frac{w_2}{w_1}\right)$,
where $10\log(\cdot)$ is applied to make the numerical range similar to relative positions. Then, the final positional embeddings are calculated as
\begin{align}
    p_a^{(u,v)} &= \mathrm{PE}(\tilde{p}_a^{(u,v)}), \nonumber\\
    p_b^{(u,v)} &= \mathrm{Linear}\bigg(\mathrm{Concat}\left(\mathrm{PE}(\tilde{p}_b^{(u,v)}),~\mathrm{PE}(s)\right)\bigg),
\end{align}
where $\mathrm{PE}$ is the sine-cosine positional encoding proposed by \cite{vaswani2017attention}. For $x_b$, we concatenate the relative positions and scale changes together, and use a linear layer to fit the dimension.

%%%%%%%%%%%%%%%%%%%%%%%%%%%%%%%%%%%%%%%%%%%%%%%%%%%%%%%%%%%%%%%%%%%%%%%%%%%%%%%%%%%%%%%%%%%%%%%%%%%%%%%%%%%%%%%%%%%%%%%%%%%%%%%%%%%%%%%%%%%%%%%%%%

\subsection{Loss Function}
Loss functions guide the training direction, and thus shape the characteristic of the learned representations. ID methods usually adopt a loss over the globally averaged feature to separate representations among images, while MIM employs a dense loss on image patches to learn representations within each individual image. 
Interestingly, we find that a dense loss only is enough to train both semantic alignment and spatial sensitivity well. As a result, we only employ a dense loss for training \name{}.

Once the prediction $y_b$ and target $z_b$ have been calculated, we adopt a dense loss function for each predicted token. UniGrad~\cite{tao2021exploring} is employed because it is a unified loss of ID methods and is also memory-friendly. 
To apply UniGrad on the dense level, we treat each token representation as an independent sample, \ie $y_b^i, i=1,\dots,N$. The corresponding positive sample is therefore $z_b^i$, and the negative samples consist of all the tokens from the target branch. The dense loss can then be computed according to\footnote{Here we remove the L2 normalization in the original UniGrad formulation, as we empirically find that not applying normalization to the online prediction can improve the performance (see Section~\ref{subsec:ablation}).}
\begin{equation}
\label{equ:contrastive_loss}
    L=\mathbb{E}_{\{y_b^i,z_b^i\}}\bigg[-||y_b^i - z_b^i||^2 + \lambda\sum_{u\in\mathcal{N}}(u^Ty_b^i)^2\bigg],
\end{equation}
where $y_b^i$ comes from the online prediction, its target $z_b^i$ is the positive sample, and all token representations from the target branch constitute the negative sample set $\mathcal{N}$. 
Note that most ID methods use the InfoNCE loss~\cite{van2018representation}, which will require $\mathcal{O}(|\mathcal{N}|)$ memory to calculate the similarities. This is infeasible for the dense loss because of the vast number of negative sample patches. In contrast, UniGrad~\cite{tao2021exploring} only consumes $\mathcal{O}(D^2)$ memory by first calculating the covariance matrix of negative samples.

%%%%%%%%%%%%%%%%%%%%%%%%%%%%%%%%%%%%%%%%%%%%%%%%%%%%%%%%%%%%%%%%%%%%%%%%%%%%%%%%%%%%%%%%%%%%%%%%%%%%%%%%%%%%%%%%%%%%%%%%

\vspace{-0.16em}
\subsection{Discussion}
\vspace{-0.14em}
As illustrated in Fig.~\ref{fig:compare}, we would like to further emphasize the differences between \name\ and previous works from three perspectives:

\vspace{0.3em}\noindent
(1) Compared with MIM methods, \name\ reveals that it's possible to reconstruct another augmented view rather than the same view. Such reconstruction can greatly enhance the semantic alignment of the model. We also show that strong augmentations can benefit this learning process;

\vspace{0.3em}\noindent
(2) Compared with ID methods, \name\ shows that dense supervision can greatly improve spatial sensitivity, and the model can also learn semantic alignment well with a dense loss. By employing the relative positions between two views, \name\ is able to achieve strict spatial alignment and build dense correspondence without anbiguity. We also demonstrate a considerable boost on a strong detection baseline;

\vspace{0.3em}\noindent
(3) For combining the best of MIM and ID methods, recent works have also made some attempts~\cite{zhou2021ibot,wang2022repre}. Nevertheless, their efforts do not jump out of the default setting of both frameworks, \ie they use different views only for ID, and apply MIM only on each view independently. These methods rely on both the global and dense loss as training objectives. 
In comparison, our work can naturally take the best of ID and MIM. \name\ enforces the similarity between different views from the dense level. Benefiting from our modeling, we can reveal the most important factors that influence the linear probing and object detection performances by gradually modifying ID or MIM methods to \name{} (see Section~\ref{subsec:ablation}).
\section{Experiments}
\label{sec:exp}
\subsection{Implementation Details}
ViT-B/16~\cite{dosovitskiy2020image} is used as the backbone. Transformer encoder blocks~\cite{vaswani2017attention} with BatchNorm~\cite{ioffe2015batch} are adopted as the projector and decoder. Before calculating loss, if not specified, we follow MAE~\cite{he2021masked} to apply LayerNorm without affine parameters to target, and no normalization to prediction. We set $\lambda=0.02$ in the loss function. During pre-training, we adopt the standard augmentation used in MoCo-v3~\cite{chen2021empirical}. For masking strategy, if not specified, we follow BEiT~\cite{bao2021beit} to use blockwise masking.
We evaluate our model in various downstream tasks, including full data fine-tuning, few-shot learning and linear probing on ImageNet~\cite{deng2009imagenet}, object detection on COCO~\cite{lin2014microsoft} and LVIS~\cite{gupta2019lvis}, semantic segmentation on ADE20k~\cite{zhou2017scene}, as well as several robustness benchmarks~\cite{hendrycks2021natural,hendrycks2021many,hendrycks2019benchmarking,wang2019learning}.
Please refer to Appendix~\ref{sec:appendix_details} for more implementation details.

\subsection{Main Results}
\label{subsec:main results}

\setlength{\tabcolsep}{4pt}
\begin{table*}[t]
    \small
    \begin{subtable}[t]{0.5\textwidth}
        \centering
        \resizebox{0.9\linewidth}{!}{
        \begin{tabular}{lcccc}
        \toprule
             \multirow{2}{*}{Method} & \multirow{2}{*}{Epochs} & \multicolumn{3}{c}{ImageNet} \\
             & & FT & LIN & FT$_{1\%}$ \\
        \midrule
             \demph{Supervised} & \demph{300} & \demph{81.8} & \demph{-} & \demph{-} \\
             \demph{DINO$^*$} & \demph{800$^\dagger$} & \demph{82.8} & \demph{78.2} & \demph{-} \\
             \demph{iBOT$^*$} & \demph{800$^\dagger$} & \demph{84.0} & \demph{79.5} & \demph{-} \\
             MoCo-v3 & 600$^\dagger$ & 83.0 & \default{76.7} & \default{63.4} \\
             BEiT & 800 & 83.2 & - & - \\
             MAE  & 400 & 83.1 & 62.5 & - \\
             MAE  & 1600 & \default{83.6} & 68.0 & 51.1 \\
        \midrule
            \name{}  & 400 & 83.7 & 76.8 & 61.8 \\
            \textbf{\name{}}  & 1600 & \textbf{\res{84.1}{+0.5}} & \textbf{\res{78.0}{+1.3}} & \textbf{\res{65.1}{+1.7}} \\
        \bottomrule
        \end{tabular}}
        \caption{Image classification.}
        \vspace{0.2em}
        \label{tab:classification}
    \end{subtable}
    \hfill
    \begin{subtable}[t]{0.5\textwidth}
        \centering
        \resizebox{0.9\linewidth}{!}{
        \begin{tabular}{lcccc}
        \toprule
             \multirow{2}{*}{Method} & \multirow{2}{*}{Epochs} & \multicolumn{2}{c}{COCO} & ADE20k \\
             & & AP$^{\mathrm{b}}$ & AP$^{\mathrm{m}}$ & mIoU \\
        \midrule
             \demph{Supervised} & \demph{300} & \demph{47.9} & \demph{42.9} & \demph{47.4} \\
             \demph{DINO$^*$} & \demph{800$^\dagger$} & \demph{50.1} & \demph{43.4} & \demph{46.8} \\
             \demph{iBOT$^*$} & \demph{800$^\dagger$} & \demph{51.2} & \demph{44.2} & \demph{50.0} \\
             MoCo-v3 & 600$^\dagger$ & 47.9 & 42.7 & 47.3 \\
             BEiT & 800 & 49.8 & 44.4 & 47.1\\
             MAE  & 400 & 50.6 & 45.1 & 45.0\\
             MAE  & 1600 & \default{51.6} & \default{45.9} & \default{48.1}\\
        \midrule
            \name{}  & 400 & 50.7 & 44.9 & 49.6\\
            \textbf{\name{}}  & 1600 & \textbf{\res{52.1}{+0.5}} & \textbf{\res{46.2}{+0.3}} & \textbf{\res{51.1}{+3.0}}\\
        \bottomrule
        \end{tabular}}
        \caption{Common object detection and semantic segmentation.}
        \vspace{0.2em}
        \label{tab:coco-detection}
    \end{subtable}
    \begin{subtable}{0.5\textwidth}
        \centering
        \resizebox{0.82\linewidth}{!}{
        \begin{tabular}{lccccc}
        \toprule
             Method & Epochs & AP$^{\mathrm{b}}$ & AP$^{\mathrm{b}}_{\mathrm{rare}}$ & AP$^{\mathrm{m}}$ & AP$^{\mathrm{m}}_{\mathrm{rare}}$  \\
        \midrule
             \demph{Supervised} & \demph{300} & \demph{37.2} & \demph{-} & \demph{34.9} & \demph{26.4} \\
             MoCo-v3 & 600$^\dagger$ & 37.3 & 25.5 & 35.3 & 25.8\\
             MAE  & 400 & 38.4 & 25.4 & 36.6 & 25.7 \\
             MAE  & 1600 & \default{40.1} & \default{29.3} & \default{\textbf{38.1}} & \default{29.1} \\
        \midrule
            \name{}  & 400 & 38.5 & 28.9 & 36.1 & 27.7 \\
            \makecell[l]{\textbf{\name{}}\\~} & \makecell{1600\\~} & \makecell{\textbf{40.5}\\\textbf{(\gain{+0.4})}} & \makecell{\textbf{30.9}\\\textbf{(\gain{+1.6})}} & \makecell{\textbf{38.1}\\\textbf{(\gain{+0.0})}} & \makecell{\textbf{30.1}\\\textbf{(\gain{+1.0})}} \\
        \bottomrule
        \end{tabular}}
        \caption{Long-tail object detection on LVIS.}
        \label{tab:lvis-detection}
    \end{subtable}
    \hfill
    \begin{subtable}{0.5\textwidth}
        \centering
        \vspace{-0.1cm}
        \resizebox{0.87\linewidth}{!}{
        \begin{tabular}{lcccccc}
        \toprule
             Method & Epochs & \makecell{IN-A\\top-1} & \makecell{IN-R\\top-1} & \makecell{IN-Sketch\\top-1} & \makecell{IN-C\\1-mCE} \\
        \midrule
            \demph{MSN$^*$} & \demph{1200$^\dagger$} & \demph{37.5} & \demph{50.0} & \demph{36.3} & \demph{53.4} \\
             MoCo-v3 & 600$^\dagger$ & 32.4 & \default{49.8} & \default{35.9} & \default{55.4} \\
             MAE & 1600 & \default{35.9} & 48.3 & 34.5 & 48.3 \\
        \midrule
            \name{} & 400 & 38.6 & 51.6 & 37.7 & 55.9 \\
            \makecell[l]{\textbf{\name{}}\\~} & \makecell{1600\\~} & \makecell{\textbf{43.8}\\\textbf{(\gain{+7.9})}} & \makecell{\textbf{52.5}\\\textbf{(\gain{+2.7})}} & \makecell{\textbf{38.3}\\\textbf{(\gain{+2.4})}} & \makecell{\textbf{57.1}\\\textbf{(\gain{+1.7})}} \\
        \bottomrule
        \end{tabular}}
        \caption{Robustness evaluation.}
        \label{tab:robustness}
    \end{subtable}
    \vspace{-1.5em}
    \caption{Results on downstream tasks. The numbers in gray cell are the main baselines that we compare. The bold numbers are the best results. ``FT" denotes full data finetuning. ``LIN" denotes linear probing. ``FT$_{1\%}$" denotes 1\% data finetuning. $^*$These methods use multi-crop during pretraining. $^\dagger$These methods use a symmetric loss, so the effective number of epoch should be doubled.}
    \vspace{-1.1em}
\end{table*}

\noindent
\textbf{Image Classification.} 
Tab.~\ref{tab:classification} shows the results of image classification tasks on ImageNet~\cite{deng2009imagenet}. For full data finetuning, \name\ surpasses both MIM and ID methods. For linear probing, our work outperforms MAE~\cite{he2021masked} by 10 points, and MoCo-v3~\cite{chen2021empirical} by 1.3 poins. When only 1\% data is available, \name\ can outperform MoCo-v3 by 1.7 points and MAE by 14.0 points. This validates that \name\ has obtained good semantic alignment. Moreover, \name\ can already deliver comparable results with previous works with only 400 epochs' pretraining.

\vspace{0.5em}\noindent
\textbf{Common Object Detection.} 
Tab.~\ref{tab:coco-detection} reports the performance on COCO~\cite{lin2014microsoft} detection. Compared to MoCo-v3, our method obtains 4.2 points improvement. \name\ is also better than MIM methods~\cite{bao2021beit,he2021masked}. This comparison validates that our method gets good spatial sensitivity.

\vspace{0.5em}\noindent
\textbf{Semantic Segmentation.}
Tab.~\ref{tab:coco-detection} also demonstrates segmentation results on ADE20k~\cite{zhou2017scene}. \name\ can surpass all pure ID and MIM methods over 3.0 points. Moreover, our method obtains 49.6 mIoU with only 400 epochs' pretraining, already on par with previous methods. Different from data-balanced ImageNet and COCO datasets, ADE20k contains classes that do not have enough labels. It thus demands both semantic alignment and spatial sensitivity of high quality. This shows the superiority of our method.

\vspace{0.5em}\noindent
\textbf{Long-tail Object Detection.}
Tab.~\ref{tab:lvis-detection} compares the results on LVIS~\cite{gupta2019lvis} object detection. \name\ performs on par with MAE~\cite{he2021masked} on overall AP metric, and delivers 1.6 point gain on rare classes. Different from common object detection, long-tail object detection poses higher demand for semantic alignment because of rare classes. \name\ therefore displays larger improvement.

\vspace{0.5em}\noindent
\textbf{Robustness Evaluation.}
Tab.~\ref{tab:robustness} shows the robustness evaluation on four datasets~\cite{hendrycks2021natural,hendrycks2021many,hendrycks2019benchmarking,wang2019learning}. Compared with MoCo-v3~\cite{chen2021exploring}, our method can bring an average of 4.5 gain. Compared with MAE~\cite{he2021masked}, \name\ can leads by a large margin of an average of 6.1 points. The results suggest that \name\ helps to improve the robustness of representations both over ID and MIM methods.

\setlength{\tabcolsep}{4pt}
\begin{table*}[t]
    \centering
    \small
    \resizebox{0.82\linewidth}{!}{
    \begin{tabular}{ccccccrcccc}
    \toprule
         & \makecell[c]{target\\type} & \makecell[c]{different\\views} & \makecell[c]{color\\aug} & \makecell[c]{mask\\type} & \makecell[c]{BN/LN in\\proj \& dec} & \makecell[c]{loss norm$^*$} & \makecell[c]{loss\\type} & LIN & AP$^{\mathrm{b}}$ & AP$^{\mathrm{m}}$ \\
    \midrule
    \multicolumn{10}{l}{\textit{single view with dense loss:}} \\ 
        \demph{MAE} & \demph{pixel} &  &  & \demph{random} & \demph{LN} & \demph{MAE-like} & \demph{dense} & \demph{62.5} & \demph{46.8} & \demph{42.0} \\
        (a) & pixel &  & & random & LN & MAE-like & dense & 62.3 & 47.3 & 42.5 \\
        (b) & feature &  & & random & LN & MoCo-like & dense & 48.7 & 43.5 & 39.2 \\
        (c) & pixel &  & \checkmark & random & LN & MAE-like & dense & 59.9 & 46.3 & 41.8 \\
    \midrule
    \multicolumn{10}{l}{\textit{multiple views with dense loss:}} \\ 
        (d) & pixel & \checkmark & & random & LN & MAE-like & dense & 46.2 & 38.1 & 34.8 \\
        (e) & feature & \checkmark & & random & LN & MoCo-like & dense & 69.6 & 48.5 & 43.4 \\
        (f) & feature & \checkmark & \checkmark & random & LN & MoCo-like & dense & 73.1 & 47.9 & 43.2 \\
        (g) & feature & \checkmark & \checkmark & random & BN & MoCo-like & dense & 73.6 & 48.7 & 43.7 \\
        (h) & feature & \checkmark & \checkmark & blockwise & BN & MoCo-like & dense & 74.7 & 50.0 & 44.5 \\
        \default{\textbf{(i)}} & \default{feature} & \default{\checkmark} & \default{\checkmark} & \default{blockwise} & \default{BN} & \default{MAE-like} & \default{dense} & \default{76.8} & \default{49.8} & \default{44.2} \\
    \midrule
    \multicolumn{10}{l}{\textit{multiple views with global loss:}} \\ 
        (j) & feature & \checkmark & \checkmark & random & BN & MoCo-like & global & 72.0 & 45.9 & 41.4 \\
        \demph{\makecell[c]{MoCo-v3\\with mask}} & \demph{feature} & \demph{\checkmark} & \demph{\checkmark} & \demph{random} & \demph{BN} & \demph{MoCo-like} & \demph{global} & \demph{72.2} & \demph{45.0} & \demph{40.5} \\
    \bottomrule
    \end{tabular}}
    \vspace{-0.1cm}
    \caption{Ablations on \name{}. We focus on the performances of linear probing, object detection and instance segmentation tasks. All models are pre-trained for 400 epochs. The fine-tuning length is 90 epochs for linear probing and 25 epochs for COCO detection. The line with gray cells is our final setting.$^*$MAE-like loss norm refers to apply LN without affine parameters to target and no normalization to prediction. MoCo-like loss norm refers to apply BN without affine parameters follow by $l_2$ normalization to both target and prediction.}
    \vspace{-0.4cm}
    \label{tab:ablation}
\end{table*}

\subsection{Ablation Study}
\label{subsec:ablation}
We carefully ablate the components of \name\ in this section to identify the most important factors for semantic alignment and spatial sensitivity.
We focus on the linear probing and COCO detection results. We state the key observations as follows.

\vspace{0.5em}\noindent
\textbf{Predicting Pixels or Features.} Tab.~\ref{tab:ablation}(ab) and (de) ablate what type of target to use. When the same view is used for input and target, we find that predicting raw pixels performs better than predicting features. On the contrary, it's superior to predict features if different views are used. We suspect that, for different views, predicting pixels presents a much more difficult pretext task than using the same view, whereas predicting features simplifies this reconstruction because the network can help to filter irrelevant details and extract semantic information.

\vspace{0.5em}\noindent
\textbf{Different Views.} We demonstrate the effectiveness of different views by comparing Tab.~\ref{tab:ablation}(af). Here, we choose the best-performed setting for both the same view or different views. It's shown that different views significantly improve the linear probing performance by $\sim$11 points. This justifies our claim that matching different augmented views is the key to obtain semantic alignment.

\vspace{0.5em}\noindent
\textbf{Color Augmentations.} Tab.~\ref{tab:ablation}(ac) and (ef) reports the effects of color augmentations. We observe different effects with the same view or different views. Color augmentations can help linear probing to obtain 3.5 points gain for different views, but this improvement vanishes with the same view. This coincides with the phenomena in SimCLR~\cite{chen2020simple} and MAE~\cite{he2021masked}. We presume that if the same view is adopted, the color augmentations used for the target will be leaked to the model, which spoils the color variation.

\vspace{0.5em}\noindent
\textbf{BN/LN for Projector and Decoder.} We also study how different normalizations will influence the model. The commonly used normalization in Transformer blocks is LayerNorm (LN)~\cite{ba2016layer}, while BatchNorm (BN)~\cite{ioffe2015batch} proves to be important in ID methods~\cite{he2020momentum}. We therefore try to replace LNs with BNs. Note that to preserve the vanilla ViT~\cite{dosovitskiy2020image} backbone, we only conduct this replacement for the projector and decoder. Tab.~\ref{tab:ablation}(fg) displays that BN gives slightly better results on both linear probing and dense prediction. 

\vspace{0.5em}\noindent
\textbf{Mask Type.} Tab.~\ref{tab:ablation}(gh) compares two mask strategies. It shows that blockwise mask is beneficial for both downstream tasks. We think that it should be easier to reconstruct feature by just interpolating local features. Blockwise mask masks out continuous patches, which makes this hard and forces the model to capture long-range dependency.

\vspace{0.5em}\noindent
\textbf{Loss Normalization.} Tab.~\ref{tab:ablation}(hi) ablates normalization in the loss function. MoCo-like normalization applies BN without affine parameters follow by l2 normalization to both target and prediction. MAE-like normalization only applies LN without affine parameters to the target. We find that MAE-like normalization performs better on linear probing and comparable on object detection campared with MoCo-like normalization, thus we adopt MAE-like loss normalization in our default setting.

\vspace{0.5em}\noindent
\textbf{Dense Supervision.} Finally we study the role that the dense supervision plays in Tab.~\ref{tab:ablation}(gj). By adopting the dense loss, \name\ is able to get an improvement of 2.8 points on object detection and 2.3 points on instance segmentation. This comparison validates our observation that modeling dense representations from a masked image is beneficial for dense prediction tasks.
We note that only using dense loss can also help to improve linear probing.
\section{Conclusion}
\label{sec:conclusion}

Different self-supervised learning (SSL) frameworks have their own strengths: Instance Discrimination (ID) possesses good semantic alignment, and Masked Image Modeling (MIM) has decent spatial sensitivity. In this study, we propose a new SSL framework, namely Siamese Image Modeling (\name{}), to show that it's possible to obtain two properties at the same time using only a single dense loss. We observe that (1) semantic alignment can be learned by matching different augmented views; (2) spatial sensitivity can be obtained by modeling dense representations from masked images. As a result, we propose Siamese Image Modeling (\name{}), which predicts the dense representations of an augmented view, based on another masked augmented view from the same image. \name\ is able to outperform ID and MIM methods over a wide range of downstream tasks. We hope that \name\ can bring some insights and inspirations for self-supervised pre-training, and  open new possibilities in this domain.

\vspace{0.5em}\noindent
\textbf{Limitations.} The training of \name\ is less efficient compared to that of MAE. Using fewer tokens or smaller resolutions for the target branch may reduce the computation burden. Because this paper focuses on exploring the possibility of self-supervised pretraining, \ie combining linear separability and spatial sensitivity within a single loss, and revealing the connection between ID and MIM methods, we expect to propose a more efficient way to perform \name\ pretraining in future work.

\vspace{0.5em}\noindent
\textbf{Potential negative societal impacts.} Our method has similar problems of the SSL paradigm. It requires huge computational resources to conduct large scale pretraining, which may consume a lot of electricity. Furthermore, it may possess biases in its digested data, and therefore should be used with caution.
\vspace{-1.0em}
\paragraph{Acknowledgments} The work is partially supported by the National Natural Science Foundation of China under Grants No.U19B2044, No.61836011 and No.62022048.

%%%%%%%%% REFERENCES
{\small
\bibliographystyle{ieee_fullname}
\bibliography{texts/reference}
}

\newpage
\appendix
\section{Implementation Details}
\label{sec:appendix_details}

\subsection{Pre-training}
\label{subsec:pretrain}
For pre-training, we mainly follow the setting of MAE~\cite{he2021masked}. Detailed hyper-parameters for pre-training are listed in Tab.~\ref{tab:hypers_pretrain}.

\noindent\textbf{Augmentation} We use the strong augmentations from MoCo-v3~\cite{chen2021empirical}, including random resized cropping, horizontal flipping, color jittering, grayscale conversion, Gaussian blurring and solarization. For masking strategy, we use follow BEiT~\cite{bao2021beit} to use blockwise masking~\cite{bao2021beit} with a masking ratio of 60\%. 

\noindent\textbf{Architecture.} We use the standard ViT-B/16~\cite{dosovitskiy2020image} as the backbone for both online and target branches. We stack $2$ and $4$ Transformer encoder blocks with BatchNorm~\cite{ioffe2015batch} as the projector and the decoder, respectively. Both the projector and decoder have $768$ embedding dimension and $12$ heads for each block. The EMA coefficient for target momentum encoder is initialized as $0.995$ and is applied with a cosine schedule from $0.995$ to $1.0$. Before calculating loss, we follow MAE~\cite{he2021masked} to apply LayerNorm without affine parameters to target, and no normalization to prediction.

\begin{table}[h]
    \centering
    \small
    \vspace{-0.5em}
    \caption{Hyper-parameters for pre-training.}
    \begin{tabular}{lc}
    \toprule
        Hyper-parameters & Value \\
    \midrule
        Layers & 12\\
        Hidden size & 768 \\
        FFN inner hidden size & 3072 \\
        Attention heads & 12 \\
        Patch size & $16 \times 16$ \\
    \midrule
        Data augment & \makecell{RandomResizedCrop\\RandomHorizontalFlip\\ColorJitter\\RandomGrayscale\\GaussianBlur\\Solarize} \\
        Mask strategy & Blockwise mask \\
        Mask ratio & 60\% \\
        Input resolution & $224\times 224$ \\
    \midrule
        Training epochs & 1600 \\
        Batch size & 4096 \\
        Adam $\beta$ & (0.9, 0.95) \\
        Peak learning rate & $1.0\times 10^{-3}$\\
        Learning rate schedule & cosine \\
        Warmup epochs & 40 \\
        Weight decay & 0.05 \\
        EMA coeff & 0.995 \\
        EMA schedule & cosine \\
    \bottomrule
    \end{tabular}
     \vspace{-0.5em}
    \label{tab:hypers_pretrain}
\end{table}

\subsection{Finetuning with 100\% Data}
\label{subsec:ft}
We follow the finetuning setting of MAE~\cite{he2021masked} except that we search for the optimal learning rate. Other hyper-parameters are listed in Tab.~\ref{tab:hypers_finetune}.
\begin{table}[h]
    \centering
    \small
    \caption{Hyper-parameters for ImageNet finetuning.}
    \vspace{0.5em}
    \begin{tabular}{lc}
    \toprule
        Hyper-parameters & Value \\
    \midrule
        Layers & 12\\
        Hidden size & 768 \\
        FFN inner hidden size & 3072 \\
        Attention heads & 12 \\
        Patch size & $16 \times 16$ \\
        Layer-wise learning rate decay & 0.65 \\
    \midrule
        Erasing prob. & 0.25 \\
        Rand augment & 9/0.5 \\
        Mixup prob. & 0.8 \\
        Cutmix prob. & 1.0 \\
        Input resolution & $224\times 224$ \\
    \midrule
        Finetuning epochs & 100 \\
        Batch size & 1024 \\
        Adam $\beta$ & (0.9, 0.999) \\
        Peak learning rate & $1.0\times 10^{-3}$\\
        Learning rate schedule & cosine \\
        Warmup epochs & 5 \\
        Weight decay & 0.05 \\
    \midrule
        Label smoothing & 0.1 \\
        Stock. depth & 0.1 \\
    \bottomrule
    \end{tabular}
    \label{tab:hypers_finetune}
\end{table}

\subsection{Linear Probing}
We follow the linear probing setting of MAE~\cite{he2021masked} while always  searching for the optimal learning rate. Specifically, an extra BatchNorm layer without affine transformation is added before the final linear classifier. Other hyper-parameters are listed in Tab.~\ref{tab:hypers_linear}.
\begin{table}[h]
    \centering
    \small
    \caption{Hyper-parameters for ImageNet linear probing.}
    \vspace{0.5em}
    \begin{tabular}{lc}
    \toprule
        Hyper-parameters & Value \\
    \midrule
        Layers & 12\\
        Hidden size & 768 \\
        FFN inner hidden size & 3072 \\
        Attention heads & 12 \\
        Patch size & $16 \times 16$ \\
    \midrule
        Data augment & \makecell{RandomResizedCrop\\RandomHorizontalFlip} \\
        Input resolution & $224\times 224$ \\
    \midrule
        Training epochs & 90 \\
        Batch size & 16384 \\
        Optimizer & LARS \\
        Peak learning rate & 3.2 \\
        Learning rate schedule & cosine \\
        Warmup epochs & 10 \\
        Weight decay & 0.0 \\
    \bottomrule
    \end{tabular}
    \label{tab:hypers_linear}
\end{table}

\subsection{Finetuning with 1\% Data}
For few-shot evaluation, we follow the practice in \cite{assran2022masked}. Specifically, we freeze the backbone and extract representations for each image. Then the cyanure package~\cite{mairal2019cyanure} is used to apply $\l_2$-regularized logistic regression on the representations. Note that for MAE, we report partial finetuning result because it is better than just training a linear classifier~\cite{assran2022masked}.

\subsection{COCO Detection}
We follow \cite{li2022exploring} to evaluate on COCO~\cite{lin2014microsoft}. We adjust the learning rate schedule so as to drop the learning rate once the performance saturates. Hyper-parameters are listed in Tab.~\ref{tab:hypers_coco}.
\begin{table}[h]
    \centering
    \small
    \caption{Hyper-parameters for COCO detection.}
    \vspace{0.5em}
    \begin{tabular}{lc}
    \toprule
        Hyper-parameters & Value \\
    \midrule
        Layers & 12\\
        Hidden size & 768 \\
        FFN inner hidden size & 3072 \\
        Attention heads & 12 \\
        Patch size & $16 \times 16$ \\
        Layer-wise learning rate decay & 0.7 \\
    \midrule
        Data augment & large scale jittor \\
        Input resolution & $1024\times 1024$ \\
    \midrule
        Finetuning epochs & 100 \\
        Batch size & 64 \\
        Adam $\beta$ & (0.9, 0.999) \\
        Peak learning rate & $1.0\times 10^{-4}$\\
        Learning rate schedule & step \\
        Warmup length & 250 iters \\
        Weight decay & 0.1 \\
    \midrule
        Stock. depth & 0.1 \\
    \midrule
        Relative positional embeddings & \checkmark \\
    \bottomrule
    \end{tabular}
    \label{tab:hypers_coco}
\end{table}

\subsection{Semantic Segmentation}
We follow \cite{bao2021beit} to use UperNet~\cite{xiao2018unified} as the segmentation network. We use the open-source code from mmsegmentation~\cite{mmseg2020} and only change pretrained backbone. Hyper-parameters are listed in Tab.~\ref{tab:hypers_seg}.
\begin{table}[h]
    \centering
    \small
    \caption{Hyper-parameters for ADE20k semantic segmentatioin.}
    \vspace{0.5em}
    \begin{tabular}{lc}
    \toprule
        Hyper-parameters & Value \\
    \midrule
        Layers & 12\\
        Hidden size & 768 \\
        FFN inner hidden size & 3072 \\
        Attention heads & 12 \\
        Patch size & $16 \times 16$ \\
        Layer-wise learning rate decay & 0.65 \\
    \midrule
        Data augment & \makecell{RandomCrop\\RandomFlip\\PhotoMetricDistortion} \\
        Input resolution & $512\times 512$ \\
    \midrule
        Finetuning length & 160k iters \\
        Batch size & 16 \\
        Adam $\beta$ & (0.9, 0.999) \\
        Peak learning rate & $1.0\times 10^{-4}$\\
        Learning rate schedule & linear \\
        Warmup length & 1500 iters \\
        Weight decay & 0.05 \\
    \midrule
        Stock. depth & 0.1 \\
    \midrule
        Relative positional embeddings & \checkmark \\
    \bottomrule
    \end{tabular}
    \label{tab:hypers_seg}
\end{table}

\begin{table}[h]
    \centering
    \small
    \caption{Hyper-parameters for LVIS detection.}
    \vspace{0.5em}
    \begin{tabular}{lc}
    \toprule
        Hyper-parameters & Value \\
    \midrule
        Layers & 12\\
        Hidden size & 768 \\
        FFN inner hidden size & 3072 \\
        Attention heads & 12 \\
        Patch size & $16 \times 16$ \\
        Layer-wise learning rate decay & 0.7 \\
    \midrule
        Data augment & large scale jittor \\
        Input resolution & $1024\times 1024$ \\
    \midrule
        Finetuning epochs & 100 \\
        Batch size & 64 \\
        Adam $\beta$ & (0.9, 0.999) \\
        Peak learning rate & $2.0\times 10^{-4}$\\
        Learning rate schedule & step \\
        Warmup length & 250 iters \\
        Weight decay & 0.1 \\
    \midrule
        Stock. depth & 0.1 \\
    \midrule
        Relative positional embeddings & \checkmark \\
    \bottomrule
    \end{tabular}
    \label{tab:hypers_lvis}
\end{table}
\subsection{LVIS Detection}
We follow \cite{li2022exploring} to evaluate on LVIS~\cite{gupta2019lvis}. We adjust the learning rate schedule so as to drop the learning rate once the performance saturates. Hyper-parameters are listed in Tab.~\ref{tab:hypers_lvis}.

\subsection{Robustness benchmarks}
We finetune the model on original ImageNet using the setting in Tab.~\ref{tab:hypers_finetune}, and test it on different validation sets without further finetuning.

\section{Attribution of Assets}
ImageNet is subject to the ImageNet terms of access~\cite{imagenet_terms}. COCO 2017 is publicly available under the Creative Commons Attribution 4.0 License. As far as we know, they do not contain any personally identifiable information or offensive content.

\end{document}